\documentclass[3p,times]{elsarticle}

\usepackage{graphics}
\usepackage{lineno}
\usepackage{amsmath}
\usepackage{makecell, multirow, diagbox}
\usepackage{url}

\usepackage{CJKutf8}

\usepackage{caption}
 \usepackage{booktabs}
\usepackage{algorithm}
\usepackage{algpseudocode}

\newtheorem{definition}{Definition}
\newtheorem{example}{Example}

\newdefinition{remark}{Remark}
\newproof{proof}{Proof}

\usepackage[
CJKbookmarks=true,
bookmarksnumbered=true,
bookmarksopen=true,
colorlinks,
pdfborder=001,
linkcolor=blue,
anchorcolor=blue,
citecolor=blue,
urlcolor=blue,
]{hyperref}

\usepackage[english]{babel}
\addto\extrasenglish{%
}

\modulolinenumbers[5]

\begin{document}

\begin{frontmatter}

\title{Knowledge-guided Unsupervised Rhetorical Parsing for Text Summarization}

\author[mymainaddress,mysecondaryaddress]{Shengluan Hou\corref{mycorrespondingauthor}}
\ead{houshengluan1989@163.com}

\author[mymainaddress,lusfirstaddress]{Ruqian Lu}
\ead{rqlu@math.ac.cn}


\cortext[mycorrespondingauthor]{Corresponding author}

\address[mymainaddress]{Institute of Computing Technology, Chinese Academy of Sciences, Beijing 100190, China}
\address[mysecondaryaddress]{University of Chinese Academy of Sciences, Beijing 100049, China}
\address[lusfirstaddress]{Academy of Mathematics and Systems Sciences $\&$ Key Lab of MADIS, Chinese Academy of Sciences, Beijing 100190, China}

\begin{abstract}
Automatic text summarization (ATS) has recently achieved impressive performance thanks to recent advances in deep learning and the availability of large-scale corpora. To make the summarization results more faithful, this paper presents an unsupervised approach that combines rhetorical structure theory, deep neural model and domain knowledge concern for ATS. This architecture mainly contains three components: domain knowledge base construction based on representation learning, attentional encoder-decoder model for rhetorical parsing and subroutine-based model for text summarization. Domain knowledge can be effectively used for unsupervised rhetorical parsing thus rhetorical structure trees for each document can be derived. In the unsupervised rhetorical parsing module, the idea of translation was adopted to alleviate the problem of data scarcity. The subroutine-based summarization model purely depends on the derived rhetorical structure trees and can generate content-balanced results. To evaluate the summary results without golden standard, we proposed an unsupervised evaluation metric, whose hyper-parameters were tuned by supervised learning. Experimental results show that, on a large-scale Chinese dataset, our proposed approach can obtain comparable performances compared with existing methods.
\end{abstract}

\begin{keyword}
Automatic text summarization \sep Rhetorical structure theory \sep Domain knowledge base \sep Attentional encoder-decoder  \sep Natural language processing
\end{keyword}

\end{frontmatter}


\section{Introduction}
Automatic text summarization (ATS) aims to produce a condensed representation while keeping the salient elements from one or a group of topic-related documents, which is a potential research area receiving considerable attentions from academia to industry. With the amounts of data are being generated in the Web age, ATS plays an increasingly important role in addressing the problem of how to acquire information and knowledge in a fast, reliable and efficient way. Generally, ATS can be categorized into two types: extractive summarization and abstractive counterpart \cite{aggarwal2018text,gambhir2017recent}. Extractive text summarization approaches directly extract salient words, sentences, or other granularities of texts to produce the summary. Conversely, abstractive models paraphrase the salient contents using NLG techniques. Abstractive methods concerns the generation of new sentences, new phrases while retaining the same meaning as the same source documents have, which are more complex than extractive ones. Extractive approaches are now the mainstream ones. According to the number of input documents, ATS can also be classified into single-document summarization (SDS) and multi-document summarization (MDS) methods.

In this paper, we focus on extractive SDS task. The key to SDS is how to score the salience of candidate text summary units (i.e. sentences, clauses, ect). There are lexical chain based approaches \cite{hou2017holographic}, classical machine learning approaches \cite{ferreira2013assessing}, graph-based unsupervised methods \cite{mihalcea2004textrank} etc. With the recent advances in deep learning, ATS has benefited much from these new ideas and gained considerable improvements \cite{cheng2016neural,nallapati2017summarunner,wu2018learning,zhou2018neural}. 
These approaches are described in more detail in \autoref{sec_related_works}. To make the results more faithful and coherent, we incorporate discourse structures into summarization generation.

Discourse structure theories involve understanding the part-­whole nature of textual documents. The task of rhetorical parsing, for example, involves understanding how two text spans are related to each other in the context.
As Web mining extracts latent knowledge from the Web content \cite{yaqoob2016big}, rhetorical parsing reveals the meaningful knowledge out of vast amounts of documents and help in improving many NLP applications. The theoretical foundation is rhetorical structure theory (RST) \cite{Mann1988Rhetorical}, which a comprehensive theory of discourse organization. RST investigates how clauses, sentences and even larger text spans connect together into a whole. RST assumes that discourse is not merely a collection of random utterances but the discourse units connect to each other as a whole in a logical and topological way. 
RST explains text coherence by postulating a hierarchical, connected tree-structure (denote as RS-tree) for a given text \cite{das2014signalling}, in which every part has a role, a function to play, with respect to other parts in the text. RST has been empirically proved useful for improving the performance of NLP tasks that need to combine meanings of larger text units, such as single-document summarization \cite{hirao2013single,louis2010discourse}, QA and chabot \cite{galitsky2017chatbot} and text classification \cite{ji2017neural,kraus2019sentiment}.

Our proposed model can benefit from RST for faithful results, which mainly consists of three components: domain knowledge base construction based on representation learning, attentional encoder-decoder model for rhetorical parsing and subroutine-based model for text summarization. The first component extracts domain keywords based on representation learning. The domain keywords contain three types: acting agents, major influence factors and dynamics of a domain. The second component leverages the output of the first component for RS-tree construction, which will be fed to the third component for summary generation.

Our aim is Chinese-oriented text summarization. To alleviate the problem of data scarcity, we leverage the labeled English data RST-DT \cite{carlson2003building} and map texts of English and Chinese into the same latent space, from which the rhetorical relation between two Chinese text spans can be determined. Furthermore, the last component extracts summary texts from the derived RS-trees. The generated summary from our subroutine-based text summarization model can be always balanced between nucleus and satellite subtrees.

The contributions of this work can be concluded as follows:
\begin{itemize}
\item We first proposed an unsupervised Chinese-oriented rhetorical parsing method. Existing rhetorical parsing methods are English-oriented, supervised methods that often trained on RST-DT, a human-annotated discourse treebank of WSJ articles under the framework of RST. Our proposed method leverages the idea of translation and embeds the Chinese and English texts in the same latent space. In this way, the rhetorical relations between Chinese text spans can be determined by the the rhetorical relations in RST-DT. 
\item Domain knowledge was utilized in the rhetorical parsing procedure, which was constructed based on representation learning. Domain knowledge was used in two aspects: one for discourse segmentation and the other one for guiding rhetorical structure inference. Furthermore, attention mechanism was adopted in rhetorical parsing thus the attention weights enable our model has the ability to focus on relevant and drown-out irrelevant parts of the input. 
\item Different from the majority of literature, our subroutine-based summarization model is purely based on the generated rhetorical structure. The basic processing unit is elementary discourse unit (EDU), which is relative shorter than sentence. Thus the generated summary can be more informative. This model is based on `importance first' principle, each time the `currently' most important EDU from the rhetorical structure will be selected one by one mechanically. The `importance first' principle makes the selection of EDUs alternated between nucleus and satellite subtrees. Thus  the generated summary can always be balanced.
\item We also proposed an unsupervised summarization evaluation metric. This evaluation metric considers many aspects of how faithful a generated summary is. To make this evaluation metric more effective, the hyper-parameters were tuned by supervised learning on the golden standard of DUC2002.
\end{itemize}

The remainder of this paper is organized as follows. \autoref{sec_related_works} reviews some related works, including approaches about domain knowledge, rhetorical parsing and automatic text summarization. \autoref{dkb_construction} is domain knowledge base construction based on representation learning. The large-scale Chinese dataset and experimental results on it will also be given. The unsupervised rhetorical parsing approach is elaborated in \autoref{sec_uns_rst_parsing}, in which the idea of translation and attention mechanism were adopted. \autoref{sec_subroutine_ts} is about the subroutine-based text summarization. An unsupervised summarization evaluation metric and experimental results are shown in \autoref{sec_experiments}. The paper is concluded with a brief summary and an outlook for further research in \autoref{sec_conclusion}.

\section{Related Works}
\label{sec_related_works}
In this section, we briefly review some related works. In \autoref{subsec_domain_know}, we will first discuss works about domain knowledge. \autoref{subsec_rst} then introduces rhetorical structure theory, which is an important theoretical foundation of our work. Finally, the latest and classical approaches of automatic text summarization will be described in \autoref{subsec_ats}. 

\subsection{Domain Knowledge}
\label{subsec_domain_know}
Knowledge is power. Domain knowledge plays a significant role in many NLP tasks. For instance, the knowledge graph (KG) is a knowledge base proposed by Google to enhance its search engine's results with information gathered from a variety of sources. Li and Mao \cite{li2019knowledge} proposed an effective way of combing human knowledge and information from data for CNN to achieve better performance. They presented K-CNN: a knowledge-oriented CNN for causal relation extraction. In K-CNN, the convolutional filters are automatically generated based on WordNet and FrameNet. The data-oriented channel is used to learn other important features of causal relation from the data. Lu et al. \cite{lu2018study} studied the concepts of big knowledge, big-knowledge system and big-knowledge engineering. Ten massiveness characteristics for big knowledge and big-knowledge systems are defined and explored. Zheng \cite{zheng2018applications} explored how to enable humans to use big knowledge correctly and effectively in biomedical domain. There are also some knowledge-based text summarization methods, we refer to \cite{goldstein2016automated,timofeyev2018building}.

Domain knowledge keyword extraction is defined as the task that automatically identifies a set of the terms that best describe the domain of documents \cite{onan2016ensemble}. Generally, domain keyword extraction approaches can be divided into two categories as unsupervised methods and supervised methods. TF-IDF is one of the simplest unsupervised approaches. The top-$k$ high TF-IDF value words are chosen as keywords. Until now, TF-IDF remains a strong unsupervised baseline \cite{marujo2015automatic}. TextRank \cite{mihalcea2004textrank} is another typical unsupervised method, which formulates keyword extraction as ``recommendation''. The supervised methods often take keyword extraction as classification problems \cite{bharti2017automatic}. However, a number of annotated dataset is needed, which is limited for unlabeled data. Kong et al. \cite{kong2018construction} constructed a Chinese sentiment lexicon using representation learning. A skip-gram model was built to predict word embeddings according to the context words and their composing characters, whose outputs were then fed into a Random Forest (RF) classifier. Words of the same polarity were then grouped together to form the sentiment lexicon.

With regard to KG, YAGO is automatically extracted from Wikipedia and other sources. YAGO2 \cite{hoffart2013yago2} contains 447 million facts about 9.8 million entities, in which an article in Wikipedia becomes an entity. DBpedia \cite{lehmann2015dbpedia} extracts fact triples from 111 different language versions of Wikipedia. To tackle the problem of low recall for pattern-based approaches, Angeli et al. \cite{angeli2015leveraging} leveraged dependency parsing tree for relation triple extraction. They constructed a few patterns for canonically structured sentences, and shift the focus to a classifier which learns to extract self-contained clauses from long sentences. On the other hand, the key idea of KG embedding is to embed components of a KG into continuous vector spaces and thus to simplify the manipulation while preserving the inherent structure of the KG \cite{wang2017knowledge}. Typical methods contain TransE \cite{bordes2013translating}, TransH \cite{wang2014knowledge}, TransR \cite{lin2015learning}, etc. KG embedding has been applied to and benefits a wide variety of downstream NLP tasks such as KG completion, question answering, and so on.

\subsection{Rhetorical Structure Theory}
\label{subsec_rst}

Rhetorical structure theory \cite{Mann1988Rhetorical} is a comprehensive theory of text organization. With more and more attentions on this theory, RST has been applied to many high-level NLP applications since Marcu's earlier works on RST parsing and applications on text summarization \cite{marcu2000theory}. RST is now one of the most popular theories for discourse analysis. 

Central to RST is rhetorical relation, which exists between two neighboring text units. The interpretation of how text spans are semantically related to each other described by rhetorical relations is crucial to retrieve important information from documents.There are two types of rhetorical relations: mononuclear relations and multi-nuclear relations. In the former ones, one of the text spans is more important than the other one, which play the role of nucleus and satellite respectively. One the other hand, all text spans are equally salient in multi-nuclear relations, which are all play the role of nucleus. Nucleus and satellite play different roles to the writer's purpose. In general, what nucleus of a rhetorical relation expresses is more essential than what satellite expresses; The nucleus is comprehensible independent of the satellite, but not vice versa.

According to RST, the minimum processing unit is EDU. EDU acts as a syntactic constituent that has independent semantics. In this sense, an EDU corresponds to a clause or a simple sentence. RST explains text coherence by postulating a hierarchical, connected tree-structure (i.e. RS-tree) for a given text. In the RS-tree, each leaf node corresponds to an EDU. Each internal node corresponds to a larger text span which captures the rhetorical relation between its two children.

Rhetorical parsing aims to generate EDU sequences and RS-trees for given documents. It involves finding roles for every granularity of text spans and rhetorical relations that hold between them. There are rule-based methods, traditional machine learning methods and deep learning methods. LeThanh et al. \cite{Lethanh2004Generating} used syntactic information and cue phrases to segment sentences and integrated constraints about textual adjacency and textual organization to generate best RS-trees. Tofiloski et al. \cite{tofiloski2009syntactic} presented a syntactic rules and lexical rules based discourse segmenter (SLSeg). Soricut and Marcu's SPADE model \cite{soricut2003sentence} used two probabilistic models for sentence-level analysis, one for segmentation and the other for RS-tree building. After that, most research focused on SVM-based discourse analysis. They regarded relation identification as classification problem \cite{feng2012text,HernaultPdI10}. Joty et al. \cite{joty2013combining} first used Dynamic Conditional Random Field (DCRF) for sentence-level discourse analysis, and then proposed a two stage rhetorical parser. Recent advances in deep learning led to further progress in rhetorical parsing. DPLP \cite{ji2014representation} is a representation learning method, whose main idea is to project lexical features into a latent space. DPLP constructs RS-trees in a shift-reduce way. A multi-class linear SVM classifier was learned to decide whether shift or reduce operation would be taken. Li et al.'s recursive method \cite{li2014recursive} contains two components. The first is to obtain the distributed representation for sentences using recursive convolution based on its syntactic tree. The second component contains two classifiers, one is used for determining whether two adjacent nodes should be merged. If so, the other one selects the appropriate rhetorical relation to the new merged subtree.

\subsection{Automatic Text Summarization}
\label{subsec_ats}

Automatic text summarization has spurred a surge of research and experimentation since its remarkable effect in modern Web age. With the fast development of deep learning technologies, many efforts applied encoder-decoder models into ATS. The usage of attention mechanism into text summarization was first brought to prominence by Rush et al. \cite{rush2015neural}. This attentional encoder-decoder abstractive model was trained on large-scale Gigaword\footnote{\url{https://catalog.ldc.upenn.edu/ldc2003t05}} dataset. Its variants and further improvements include \cite{lin2018global,nallapati2017summarunner} et al. Neural extractive methods are also popular, such as pointer network-based models \cite{cheng2016neural,see2017get}, SummaaRuNNer \cite{nallapati2017summarunner}, SWAP-NET \cite{jadhav2018extractive}, etc. Most of the extractive models are trained on CNN/DM\footnote{\url{https://github.com/deepmind/rc-data}} dataset. However, large-scale dataset is necessary for these neural models since they are purely data-driven. For MDS, parallel data is scarce and costly to obtain. To tackle this predicament, Lebanoff et al. presented PG-MMR \cite{lebanoff2018adapting}, an adaptation
method from single to MDS, to generate abstract summaries from multiple documents. This method is new, but also affected by the data source and data scale.

Besides the above deep learning-based approaches, there are also other solutions, such as traditional machine learning-based methods, optimization-based methods, graph-based ones, etc. ATS was taken as an optimization problem in \cite{gillick2009scalable}, the ILP-based method did exact inference under a maximum coverage model. Other traditional machine learning-based methods take TF-IDF, n-gram, the position and others as features to extract summary sentences. For more details, we refer to \cite{aggarwal2018text,gambhir2017recent}. Graph-based methods have become increasingly prevalent and far-reaching since their easy implementation and relative good performance, such as Textrank \cite{mihalcea2004textrank}. Another representative of unsupervised algorithm is SummCoder \cite{joshi2019summcoder}, whose summary sentence selection module contains three metrics: sentence content relevance is measured by a deep auto-encoder network, sentence novelty is measured by sentence similarity based on sentence embeddings and sentence position relevance is derived by a hand-designed score function.

The authors of RST have long speculated that the nuclei in RS-tree constitute an adequate summarization of the text. It was first validated by Marcu \cite{marcu1997discourse}.
Louis et al. \cite{louis2010discourse} proved that the structure features (i.e. position in the global structure of the whole text) of RS-tree are the most useful feature to compute the salience of text spans. Hirao et al. \cite{hirao2013single} treated summary generation as a tree knapsack problem. They transformed an RS-tree into a dependency-based discourse tree (DEP-DT), which can be directly used to take tree trimming method for text summarization. For MDS, to address redundancy problem, Zahri et al. \cite{zahri2015exploiting} used RS-trees for cluster-based MDS. They utilized rhetorical relations that exist between two sentences to group similar sentences into multiple clusters to identify themes of common information, from which candidate summary sentences were extracted. In this paper we propose further contribution to this approach, focusing on unsupervised extractive summarization.

\section{Domain Knowledge Base Construction Based on Representation Learning}
\label{dkb_construction}

Domain knowledge plays a significant role in many NLP tasks. At present, most of the existing knowledge bases are in the form of knowledge graph, such as YAGO, DBpedia, etc, which generally consists of entity and relation triples. The knowledge triples in a KG are composed of two entities along with their relation, which are in the form of $<e1, r1, e2>$, where $e1$ and $e2$ are entities that often nouns or noun phrases, $r1$ is the relation between $e1$ and $e2$. However,  knowledge keywords for a domain are also indispensable. For a domain, knowledge keywords can provide a panorama for this domain. In this section, we propose a framework of constructing domain knowledge base on the basis of representation learning. Our proposed domain knowledge contains three types of keywords: acting agents, major influence factors and dynamics of a domain.

We define the domain as:

\begin{definition}[Domain]
A domain is a particular area of human knowledge. Such as education, finance, et al.
\end{definition}

For a domain, keywords can be regarded as the knowledge generalization of the full text in a corresponding literature and help readers to quickly grasp the core idea, core technique, or core methodology, etc. In general, two different domains have different knowledge keywords, but maybe with some common knowledge keywords. The definition of domain knowledge keyword is given in Definition \ref{def_dkk}.

\begin{definition}[Domain Knowledge Keyword, DKK]
 A domain knowledge keyword is a basic and characteristic element of this domain, which is represented by a word or phrase and is often referred to when talking about some aspects of this domain. DKK can generalize the main topics of domain texts.
\label{def_dkk}
\end{definition}

\begin{example}
 ``Teacher'', ``Student'', ``Professor'', ``Teach'', ``Learn'', ``Library'', ``Course'', ``Doctoral''  are DKKs of the domain ``Education''.
\end{example}

Two different domains may share some of their DKKs (e.g. ``Library'' may be a DKK of some other domains), but never share their whole sets of DKKs. The less the size of the shared DKKs, the more are the two domains different from each other.

We argue that besides nouns, verbs and adjectives (adverbs) also serve as the key components. In the above example, ``Teacher'' is a noun, ``Teach'' is a verb, and ``Doctoral'' is an adjective. In fact, these three types of keywords constitute the main types of DKKs. For each domain, we construct domain knowledge from large-scale texts. The DKB in this work is composed of a set of triples containing domain keywords. 
\begin{definition}[Domain Knowledge Base, DKB]
For a domain, the DKB can be represented by a triple:
\begin{equation}
<A, P, T>
\label{eq_nva_triple}
\end{equation}
where
\begin{itemize}
\item  $A$ denotes nouns and named entities, each of which represents the acting agents of this domain;
\item $P$ acts as the major influence factors of this domain, which are nouns;
\item  $T$ denotes the concepts about dynamics of this domain, each of which is often adjective or adverb. 
\end{itemize}
These three types of keywords constitute a full DKB for a domain.
\end{definition}

Our goal is to construct the DKB for each domain in a fast and efficient manner. Traditional methods can obtain high accuracy but with low recall, it also need much efforts when used to a new domain. On the other hand, the more and more popular word embedding methods have the pros of robust and efficient. It can be trained on large scale dataset without any other extra resource. We leverage the representation learning from word embedding methods for DKB construction.

\begin{definition}[Domain Knowledge Base Construction, DKBC]
Given a large set of documents that consists of texts for several domains $\{D_1, D_2, \cdots, D_t\}$, DKBC aims to extract a DKB from each domain texts $D_i (1\le i \le t)$, the constructed DKB is in the form of (\ref{eq_nva_triple}).
\end{definition}

Specially, for each domain $D_i$, given the corresponding documents $\{d_1, d_2, \cdots, d_k\}$, DKBC can automatically generate three types of DKKs as defined in Definition \ref{def_dkk}. All generated DKKs can constitute the DKB (denote as $DKB_i$) in the form of (\ref{eq_nva_triple}) such that:
\begin{itemize}
\item If $w_m\in DKB_i$, then $w_m \in DICT_i$, where $DICT_i$ is the vocabulary taken from $D_i$;
\item Suppose $DKB_i=<A_i, P_i, T_i>$, $w_p \in A_i, w_q \in P_i$, then $p\neq q$.
\end{itemize}

To obtain better results, our model is an integration of three different models. The first one is representation learning based model, which we call VWRank. The other two models are TF-IDF model and TextRank model. TF-IDF is an important indicator
of the word's saliency. TextRank is an ``recommendation'' strategy for voting salient words.

\subsection{The Architecture of VWRank}
Our DKBC model utilizes representation learning from word embedding approaches. We use the improved word representation learning with sememes method, called SE-WRL \cite{niu2017improved}. The sememe knowledge base they used is Hownet \cite{zhendong2006hownet}. SE-WRL provides different strategies, among which SE-WRL-SAT achieved the best performance according to their original paper. SE-WRL-SAT learns the original word embeddings for context words, but sememe embeddings for target words.

For each domain, we use SE-WRL-SAT to learn word representations. Then we define the similarity between two candidate words $cw_1=(x_{1}^1, x_{2}^1, \cdots, x_{n}^1)$ and $cw_2=(x_{1}^2, x_{2}^2, \cdots, x_{n}^2)$ as consine distance:
\begin{equation}
\begin{split}
Sim(cw_i, cw_j)&=\frac{cw_i\cdot cw_j}{||cw_i||*||cw_j||}\\
&=\frac{\sum\limits_{k=1}^n (x_{k}^1*x_{k}^2)}{\sqrt{\sum\limits_{k=1}^n (x_{k}^1)^2}*\sqrt{\sum\limits_{k=1}^n (x_{k}^2)^2}}
\end{split}
\label{eq_cw_sim}
\end{equation}
Obviously, the following two properties hold:
\begin{itemize}
\item $Sim(cw_i, cw_j)=Sim(cw_j, cw_i)$;
\item $Sim(cw_i, cw_j)\in [-1,1]$.
\end{itemize}

Motivated by TextRank, the score of candidate keyword $cw_i$ can be computed as:
\begin{equation}
Score(cw_i)=(1-d)+d*\sum\limits_{cw_j\in S(cw_i)}\frac{Sim(cw_j, cw_i)}{\sum\limits_{cw_k\in S(cw_j)}Sim(cw_j, cw_k)}Score(cw_j)
\end{equation}
where $d\in (0,1)$ is a damping factor, which has the role of integrating into the
model the probability of jumping from a given candidate word to another random candidate word. $S(cw_i)$ and $S(cw_j)$ are two sets of candidate words that $cw_i$ and $cw_j$ similar with, respectively. The similarity between two candidate keywords is computed by (\ref{eq_cw_sim}).
After several iterations, the $Score(cw_i)$ can converge to a fixed value.

\subsection{Model Integration}
Besides VWRank, in a domain, we also calculate the TF-IDF and TextRank values for candidate words in each domain. We denote the high score candidate keywords of VWRank, TF-IDF and TextRank as $C_{vw}$, $C_{ti}$ and $C_{tr}$. The final score of a candidate keyword is computed as:
\begin{equation}
Score(cw_i)=\alpha*I(C_{wv}, cw_i)+\beta*I(C_{ti}, cw_i)+\gamma*I(C_{tr}, cw_i)
\end{equation}
where $\alpha$, $\beta$ and $\gamma$ are harmonic coefficients,  $I(\cdot)$ is the indicator function such that
\begin{equation}
I(C, cw) = \left\{ \begin{array}{l}
1, \text{ if } cw \in C \\
0, \text{ else}
\end{array} \right.
\end{equation}

Then the final DKKs are composed of candidate keywords that further filtered by the value of:
\begin{equation}
p_{tt}(cw_i)=\frac{\text{The number of documents that } cw_i \text{ presents}}{\text{The number of all documents in this domain}}
\end{equation}

Finally, all selected DKKs will be organized in hierarchies by their semantic in Hownet.

\subsection{Dataset: SogouCA}

SogouCA\footnote{\url{https://www.sogou.com/labs/resource/ca.php}} is a large-scale Chinese corpus, which is crawled and provided by Sogou Labs from dozens of Chinese news websites, including news reports and reviews. 

Each document in SogouCA contains fields of  ``url'', ``docno'', ``contenttitle'', and ``content''. Leveraging ``url'' information, we can categorize documents into corresponding domains. Excerpts of ``url'' and its corresponding domain are shown in \autoref{tb_excerp_url_domain}.

\begin{table}
\caption{Excerpts of url and its corresponding domain}
   \centering
   \begin{tabular}{ll}
   \toprule
   Url & Domain \\
   \midrule
\url{http://www.xinhuanet.com/world/} & \begin{CJK}{UTF8}{gkai}
国际
\end{CJK}(World) \\
\url{http://news.china.com/zh_cn/international/} & \begin{CJK}{UTF8}{gkai}
国际
\end{CJK}(World) \\
\url{http://finance.sina.com.cn/}  & \begin{CJK}{UTF8}{gkai}
财经
\end{CJK}(Finance)\\
\url{http://sports.china.com/}  & \begin{CJK}{UTF8}{gkai}
体育
\end{CJK}(Sports)\\
\url{http://china.soufun.com/} & \begin{CJK}{UTF8}{gkai}
房产
\end{CJK}(House) \\
$\cdots$ & $\cdots$ \\
   \bottomrule
\end{tabular}
\label{tb_excerp_url_domain}
\end{table}

After that, we collected texts for 15 domains. We did preprocessing including delete empty or very short lines, ignore extreme long lines, etc. The statistics of documents in each domain are listed in \autoref{tb_num_doc_domain}, from which we can see that most of the domains contain tens of thousands of documents. 

\begin{table}
\caption{The statistics of documents in each domain}
   \centering
   \begin{tabular}{lllllllll}
   \toprule
   Domain & Sports & IT & Military & Olympic & Culture & House & Domestic & Entertainment \\
   \midrule
    \#Docs & 323,861 & 22,033  & 17,607 & 74,374 & 7,212 & 22,381 & 2,454 & 93,949 \\
    Avg. \#Sens. & 16.49 & 20.58  & 19.27 & 17.09 & 23.49 & 17.73 & 19.46 & 16.63  \\
    Avg. \#Words. & 334.65 & 493.37  & 447.82 & 385.83 & 505.17 & 229.97 & 490.63 & 372.77  \\
 \toprule 
  Domain & Auto & Finance & Lady & Health & Education & Society & World & Total \\
   \midrule
   \#Docs & 44,462 & 263,575  & 72,970 & 5,712 & 53,197 & 2,698 & 2,566 & 1,009,231  \\
   Avg. \#Sens. & 18.48 & 22.21  & 15.49 & 20.04 & 21.22 & 20.60 & 14.25 & 18.55  \\
   Avg. \#Words. & 413.14 & 492.01  & 277.30 & 332.22 & 418.76 & 419.98 & 333.31 & 391.75  \\
   \bottomrule
\end{tabular}
\label{tb_num_doc_domain}
\end{table}

Unlike English, to manipulate text at the word level, word segmentation is needed for Chinese text processing. We used HanLP \cite{hanlp} for Chinese word segmentation, part-of-speech tagging and named entity recognition (NER), which is a Chinese natural language processing tool.

\subsection{Experimental Results of DKBC}

We have finished the DKBC for 15 domains according to the above methods. The derived 15 DKBs can be used for other NLP applications. The statistics of DKB in each domain are shown in \autoref{tb_stat_dkb}.

\begin{table}
\caption{The statistics of DKB in each domain}
   \centering
   \begin{tabular}{lllllllll}
   \toprule
   Domain & Sports & IT & Military & Olympic & Culture & House & Domestic & Entertainment \\
   \midrule
    \#Agents & 2,687 & 622  & 561 & 1,706 & 179 & 308 & 126 & 1,318 \\
    \#Phenomenons & 6,162 & 1,580  & 1,186 & 3,416 & 439 & 996 & 267 & 4,422  \\
    \#Tendencies & 5,290 & 1,004  & 767 & 2,316 & 230 & 564 & 141 & 3,448  \\
 \toprule 
  Domain & Auto & Finance & Lady & Health & Education & Society & World & Total \\
   \midrule
   \#Agents & 533 & 2,856  & 637 & 56 &902 & 93 & 130 & 12,723  \\
   \#Phenomenons & 1,999 & 5,469  & 3,504 & 413 & 2,298 & 278 & 204 & 32,633 \\
   \#Tendencies & 1,440 & 3,308  & 3,011 & 247 & 1,579 & 135 & 139 &  23,619 \\
   \bottomrule
\end{tabular}
\label{tb_stat_dkb}
\end{table}

\begin{figure}[t]
\centering
\includegraphics[width=0.5\textwidth]{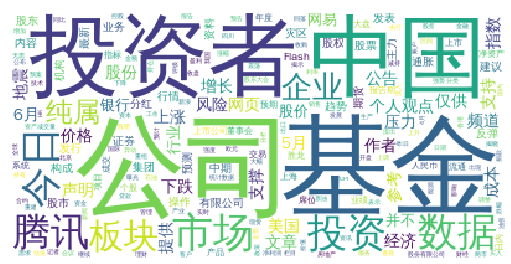}
\caption{The DKKs of ``Finance'' domain}
\label{fig_finance_wordcloud}
\end{figure}

\autoref{fig_finance_wordcloud} shows some DKKs of  ``Finance'' domain in Chinese. The DKKs contains keywords such as 
``\begin{CJK}{UTF8}{gkai}
中国
\end{CJK}(China)''(Agent), 
``\begin{CJK}{UTF8}{gkai}
投资者
\end{CJK}(investor)''(Phenomenon), 
``\begin{CJK}{UTF8}{gkai}
风险
\end{CJK}(risk)''(Phenomenon), 
``\begin{CJK}{UTF8}{gkai}
上涨
\end{CJK}(increase)''(Tendency), etc. These words can provide a panorama for domain ``Finance''.

\begin{figure}[t]
\centering
\includegraphics[width=0.5\textwidth]{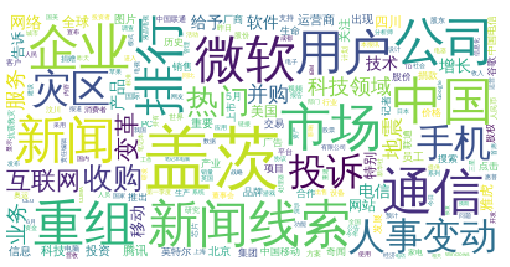}
\caption{The DKKs of ``IT'' domain}
\label{fig_it_wordcloud}
\end{figure}

On the other hand, in the domain of  ``IT'' (\autoref{fig_it_wordcloud}), some DKKs are 
``\begin{CJK}{UTF8}{gkai}
微软
\end{CJK}(Microsoft)''(Agent), 
``\begin{CJK}{UTF8}{gkai}
排行
\end{CJK}(Ranking)''(Phenomenon), 
``\begin{CJK}{UTF8}{gkai}
市场
\end{CJK}(market)''(Phenomenon),  
``\begin{CJK}{UTF8}{gkai}
上市
\end{CJK}(Be listed)''(Tendency), etc. These two examples can validate the effectiveness of our method.

\section{Unsupervised Rhetorical Parsing}
\label{sec_uns_rst_parsing}

Rhetorical structure theory was proposed as a way to attribute structure to text, which often represents a text as a tree structure. It is characterized by rhetorical relations, which reflect the semantic and functional judgments about the text spans they connect. We first give a formal definition of rhetorical structure tree.

\begin{definition}[Rhetorical Structure Tree]
Rhetorical Structure Tree (RS-tree) is a tree representation of a document under the framework of RST. The leaf nodes of a RS-tree are EDUs. Each internal node is characterized by a rhetorical relation and corresponds to a contiguous text span. The siblings are connected via a rhetorical relation such that in most cases one is nucleus and the other is satellite. The siblings are both nucleus when they are connected by a multi-nuclear relation.
\label{def_rs_tree}
\end{definition}

In Definition \autoref{def_rs_tree}, EDU is the minimal textual unit of an RS-tree, which means that it can't be split into smaller text spans. EDU acts as a syntactic constituent that has independent semantics. In this sense, an EDU functionally corresponds to a simple sentence or a clause in a complex sentence. 

\begin{definition}[Rhetorical Parsing]
Rhetorical Parsing, also called RST analysis, RST parsing, or rhetorical analysis, is a procedure of generating EDU sequences and deriving RS-trees for given texts. It involves segmenting discourse into EDUs and finding roles for every granularity of text spans (EDUs, sentences, paragraphs and even larger spans) and rhetorical relations that hold between them. 
\label{rstParsingDef}
\end{definition}

\begin{figure}[t]
\centering
\includegraphics[width=0.8\textwidth]{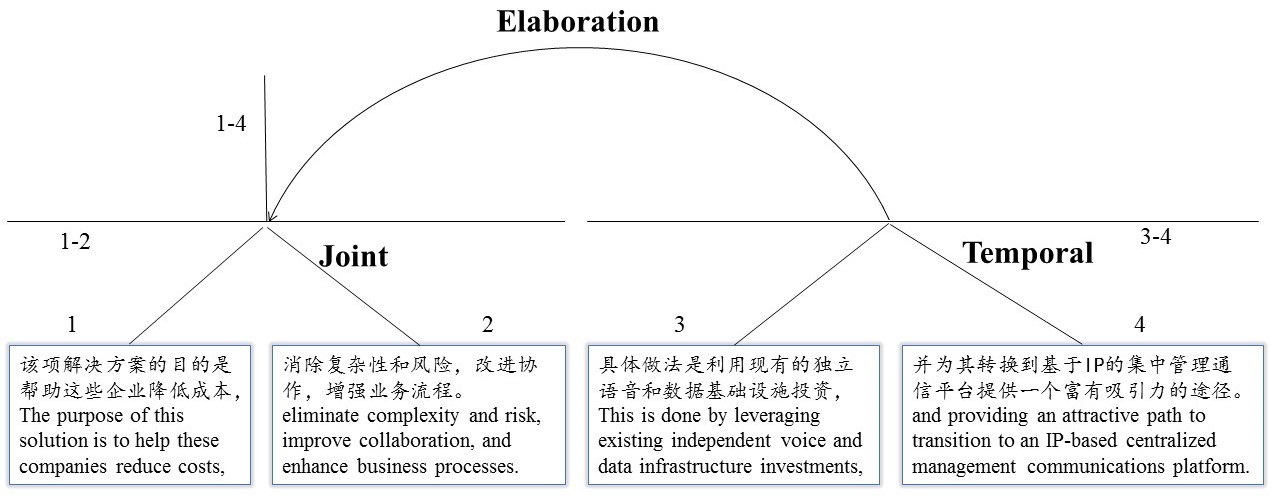}
\caption{An example of RS-tree}
\label{fig_exampleOfRSTreeFig}
\end{figure}

As depicted in Definition \autoref{rstParsingDef}, rhetorical parsing contains two steps: discourse segmentation and RS-tree construction. An example RS-tree for a given Chinese text is shown in \autoref{fig_exampleOfRSTreeFig}. The leaf nodes numbered with digits are four EDUs. The internal nodes corresponds to text spans are characterized by rhetorical relations (such as Joint and Elaboration). The arrow from $A$ to $B$ denotes $A$ and $B$ are satellite and nucleus respectively in the sense of that relation. They are both nuclei when $A$ and $B$ have multi-nuclear relation. Horizontal lines correspond to text spans, and vertical lines identify text spans which are nuclei.

\subsection{Discourse Segmentation based on Domain Knowledge Base}
Leveraging domain knowledge, we segment each document in each domain into EDU sequence. According to Definition \autoref{def_rs_tree}, EDU functionally corresponds to a simple sentence or a clause. We first segment a text into paragraphs and further sentences by punctuations. Then DKB is used for segmenting sentences into EDUs. Concretely, for a domain, given its DKB $K_d$ and domain texts $T_d$, for each text $t_i^d \in T_d$, \autoref{alg_dis_seg} is the detailed segmentation algorithm.

\begin{algorithm}[h] 
\caption{Discourse Segmentation for a Domain Text} 
\label{alg_dis_seg}
\begin{algorithmic}[1] 
\Require 
A document $t_i^d$; DKB $K_d$.
\Ensure 
EDU sequences $S_{edu}$. 
\State Segment $t_i^d$ into sentence sequence $S$ by punctuations (line break for segmenting into paragraphs, period, question mark, etc for segmenting into sentences).
\For{each sentence $s_j$ in $S$}
  \State Scan $s_j$ and match their words against the domain keywords in $K_d$;
  \State If the domain keywords of a clause have the form ``A+P+T'' or ``P+T'', then put it into $S_{edu}$.
\EndFor
\State Output the EDUs in $S_{edu}$ according to their order in the original text.
\end{algorithmic} 
\end{algorithm}

After \autoref{alg_dis_seg}, each derived EDU is a part of a sentence or clause, characterizing the domain relatedness of its elements. Moreover, most EDUs have the form of ``A+P+T'' with respect to domain keywords. For the form of ``P+T'', we borrow a agent keyword from the nearest neighbor EDU to form a complete triple. After that, each EDU has a DKB triple $<a, p, t>$.

\subsection{Rhetorical Structure Theory Discourse Treebank}

For RS-tree construction, existing models contain classical machine learning-based methods and deep learning-based methods, almost all of which are supervised methods. These approaches were trained on Rhetorical Structure Theory Discourse Treebank (RST-DT) \cite{carlson2003building}. RST-DT
was developed as a human-annotated discourse level corpus with RS-trees for 385 English-written Wall Street Journal texts. These texts were manually annotated by the professional language analysts grounded in the framework of RST. There are 78 fine-grained rhetorical relations that grouped into 18 coarse-grained relation categories. In the existing approaches, the latter 18 categories are often used for training and testing. Since there exist multi-nuclear relations, non-binary relations are often converted into a cascade of right-branching binary relations for convenience. In RST-DT, there are 21,789 EDUs and 21,404 text pairs that are characterized by rhetorical relations. 

\subsection{Attentional Encoder-decoder Model for RS-tree Construction}
\begin{figure}[t]
\centering
\includegraphics[width=0.75\textwidth]{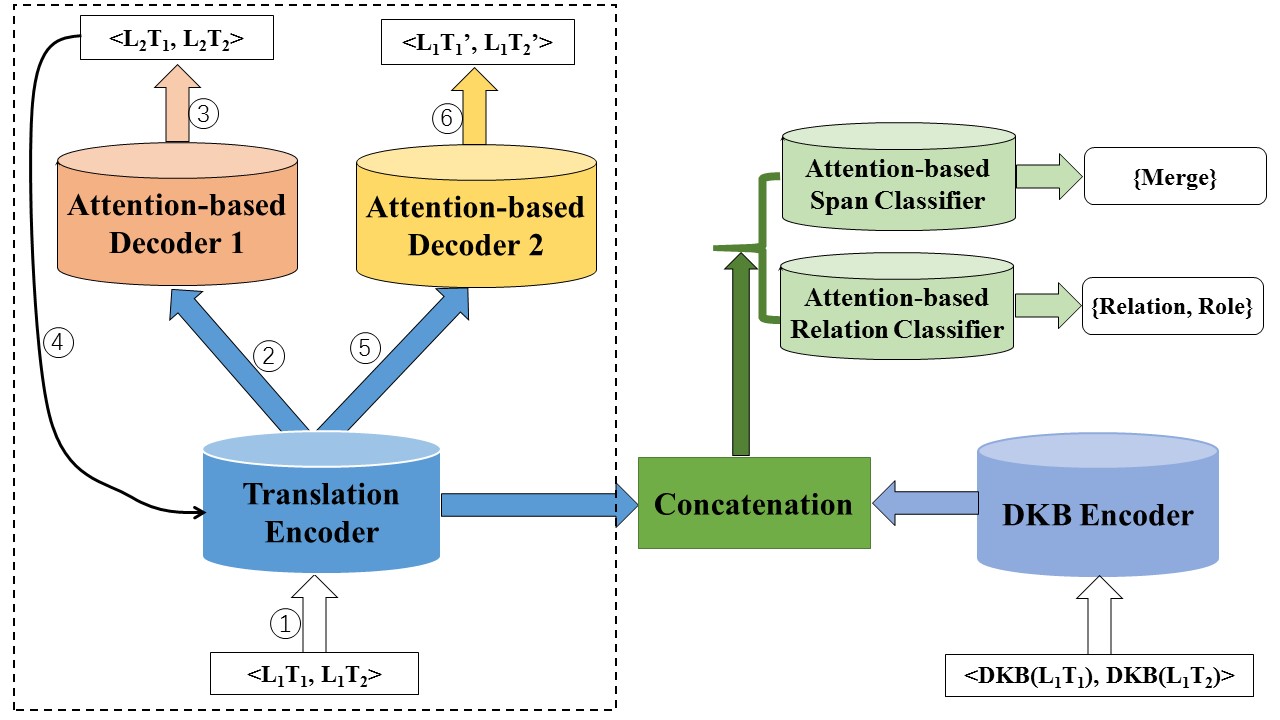}
\caption{The architecture of rhetorical relation identification}
\label{fig_archi}
\end{figure}

The objective of RS-tree construction is to find rhetorical relations between two adjacent text spans(including EDUs). Then the RS-tree can be constructed in a bottom-up way. For our Chinese-oriented rhetorical parsing work, there is no human-annotated Chinese-oriented discourse treebank like RST-DT in English. Motivated by the basic ideas of recent progress in unsupervised machine translation \cite{lample2017unsupervised}, we propose to leverage RST-DT and embed Chinese text spans and English text spans into the same latent space. Thus the rhetorical relation between two Chinese text spans can be derived by the the rhetorical relations in RST-DT. Our work is unsupervised since there's no labeled Chinese dataset is used. The architecture of rhetorical relation identification is shown in \autoref{fig_archi}.

The unsupervised rhetorical parsing model we propose is composed of two encoders, an decoder and two classifiers. The translation encoder is responsible for encoding Chinese and English texts into a latent space and the DKB encoder is used for representing domain keyword sequence. The attention-based decoder 1 and attention-based decoder 2 are the same decoder with same parameters, whose only difference is the choice of lookup tables when applying them to different languages. The two classifiers are used for rhetorical relation identification. 

In \autoref{fig_archi}, the components in dotted box are to constrain the model can map text pair from Chinese (English) to English (Chinese). Suppose $<L1T1, L1T2>$ is English (Chinese) text pair, the output of  attention-based decoder 1 is Chinese (English) $<L2T1, L2T2>$, which then will be input to the translation encoder. The output of attention-based decoder 2 is English (Chinese) $<L1T1', L1T2'>$. The object of this procedure is to learn a mapping such that translations are close in the same latent space. The translation loss function is:
\begin{equation}
L_{trans}=\sum [\Delta(<L1T1, L1T2>, <L1T1', L1T2'>)+\Delta(<L2T1, L2T2>, <L2T1', L2T2'>)]
\end{equation}
where $\Delta$ is the sum of token-level cross-entropy losses.

The second objective of our model is to train two classifiers. When $<L1T1, L1T2>$ is English text pair, we denote $<DKB(L1T1), DKB(L1T2)>$ as its domain keyword sequence. The concatenation of the two encoders' hidden states is fed into two classifiers. The attention-based span classifier is used for determining whether $L1T1$ and $L1T2$ should be merged into a new subtree and
if so the attention-based relation classifier is used to assign which relation and which role should be labeled to the merged node and its two children respectively. The loss function used for classification is also cross entropy loss.

In this work, we our propsed model is based on the sequence-to-sequence model with attention \cite{bahdanau2014neural}. These two encoders are both bidirectional-GRU which returns a sequence of hidden states whereas the decoder is also an GRU, which takes as input the previous hidden state, the current word and a context vector given by a weighted sum over the encoder states.

The final loss function is:
\begin{equation}
L=\lambda_{trans}L_{trans}+\lambda_{clas}L_{clas}
\end{equation}
where $\lambda_{trans}$ and $\lambda_{clas}$ are hyper-parameters, $L_{clas}$ is classification loss.

For inference, the input is Chinese EDU text pair along with their domain keywords and the output is (1): whether they can be merged into a subtree; (2) If so, which rhetorical relation and which role should be labeled to the merged node. Then the text of the merged node and its neighboring node's text will form the new input text pair. Loop this step until a RS-tree for a document has been constructed.

\section{Subroutine-based Model for Automatic Text Summarization}
\label{sec_subroutine_ts}

In this section, we present a subroutine-based model for automatic text summarization, which has been introduce in our previous paper \cite{lu2019arsg}. Different from the majority of literature, our subroutine-based summarization model is purely based on the generated RS-tree  from \autoref{sec_uns_rst_parsing}.

The basic processing unit is EDU, which is relative shorter than sentence. Thus the generated summary can be more informative than summary that composed of sentences. The summarization algorithm is based on `importance first' principle, each time the `currently' most important EDU from RS-tree will be selected one by one mechanically. In this way we can obtain a hierarchy of different summarizations level-wise from simple to complex by adding one more EDU at each level. There two ways of controlling the complexity of summarized result: either by specifying the word length limit or the rate of text reduction. 

When going to produce a summary, the summarization model traverses the RS-tree in a nucleus preference way. That is: (1) a nucleus node is always preferred over its sibling satellite node; (2) if node $A$ is preferred over node $B$, then all child nodes of $A$ are preferred over $B$; (3) the selection of EDUs should be alternated between the left and right subtrees of the root node whenever both subtrees are not empty. Whenever a leaf node (EDU) is traversed, the text unit represented by it will be put to the final summary. In our model, a nucleus node is always preferred over its sibling satellite node. Thus the generated summaries will always content-balanced.

All details of this subroutine-based text summarization algorithm are clarified in \autoref{alg_subroutine_ts}. The $nuc(x)$ and $sat(x)$ mean the nucleus resp. satellite child nodes of $x$. $flip (x)$ is the flip-flop function with $flip (0) = 1$ and $flip (1) = 0$. $dfp$ and $sfp$ are pointers pointing to the entry of subroutine $dfinding$ resp. $sfinding$.  $zp.$ is a formal parameter for storing a pointer. For example, after the subroutine call $dfinding (0, R (T), sat (R (T)), dfp)$ it is $zp = dfp$ in the subroutine body of $dfinding$.

\begin{algorithm}[h] 
\caption{Subroutine-based Text Summarization} 
\label{alg_subroutine_ts} 
\begin{algorithmic}[1] 
\Require 
The RS-tree $R(T)$ of a document; Summary length/cadence ratio $r$; $j = 1$; $k = 1$.
\Ensure 
The generated EDU sequences $R_{edu}$. 
\State If $R(T)$ is a leaf node then put $R(T)$ into $R_{edu}$ and goto step \ref{alg_ts_sort}, else call $dfinding (0, R(T), sat (R(T)), dfp)$.
\State Subroutine $dfinding(w, x, y, zp)$: Case: $w=0$ $\to$ If $j = 0$ then goto step \ref{alg_ts_sort}; If $nuc (x)$ is a non-leaf node then call $dfinding (w, nuc (x), y, zp)$ else put $nuc (x)$ into $R_{edu}$, call $index$ and call $zp.(flip (w), nuc (x), y, sfp)$; Case: $w=1$ $\to$ If $k = 0$ then goto step \ref{alg_ts_sort}; If $y$ is a leaf node then put $y$ into $R_{edu}$, $k:=0$, if $j=0$ then goto step \ref{alg_ts_sort} else call $index$, call $zp.(flip (w), x, y, sfp)$, else
If $nuc(y)$ is non-leaf node then call $dfinding (w, x, nuc (y), zp)$ else put $nuc(y)$ into $R_{edu}$, call $index$, call $zp.(flip (w), x, nuc(y), sfp)$.
\State Subroutine $ufinding(w, x, y, zp)$: Case: $w=0$ $\to$ If $j = 0$ then goto step \ref{alg_ts_sort} else call $zp.(flip (w), parent(x), y, sfp)$;
Case: $w=1$ $\to$ If $k = 0$ then goto step \ref{alg_ts_sort} else call $zp.(flip (w), x, parent(y), sfp)$.
\State Subroutine $sfinding(w, x, y, zp)$:
Case: $w=0$ $\to$ If $j = 0$ then call $sfinding(flip (w), x, y, sfp)$;
If $parent(x)=R(T)$ then $j:=0$, if $k=0$ then goto step \ref{alg_ts_sort} else call $sfinding(flip(w), x, y, sfp)$;
If $sibling (x)$ has been travelled then call $ufinding(w, x, y, sfp)$ else if $sibling (x)$ is a leaf node then put $sibling (x)$ into $R_{edu}$, call $index$, call $ufinding (w, x, y, sfp)$, else call $dfinding(w, sibling (x), y, sfp)$;
Case: $w=1$ $\to$ If $k = 0$ then call $sfinding(flip (w), x, y, sfp)$; 
If $parent(y)=R(T)$ then $k:=0$, if $j=0$ then goto step \ref{alg_ts_sort} else call $sfinding(flip(w), x, y, sfp)$;
If $sibling (y)$ has been travelled then call $ufinding(w, x, y, sfp)$ else 
If $sibling (y)$ is a leaf node then put $sibling (y)$ into $R_{edu}$, call $index$, call $ufinding (w, x, y, sfp)$, else call $dfinding(w, x, sibling (y), sfp)$.
\State Subroutine $index$: If the word length of $R_{edu}$ satisfies $r$ then goto step \ref{alg_ts_sort}.
\State Sort the EDUs in $R_{edu}$ according to their order in the original text.
\label{alg_ts_sort} \\ 
\Return $R_{edu}$.
\end{algorithmic} 
\end{algorithm}

\section{Experimental Results}
\label{sec_experiments}

\subsection{Training Details about Unsupervised Rhetorical Parsing}
The training of our unsupervised rhetorical parsing was carried out on SogouCA and RST-DT datasets.
We used a mini-batch stochastic gradient descent (SGD) algorithm together with Adam \cite{ruder2016overview} with initial learning rate 0.001 to train this model. In each epoch, the training data in each batch are the mixture of Chinese and English text pairs. We used Textrank for DKBC of RST-DT. The size of word embedding for both language and GRU hidden state dimensions are set to 100 and 300 respectively. For two decoders, texts are generated using greedy decoding.

\subsection{Unsupervised Quantitative Evaluation Metric}

The commonly used evaluation metric for text summarization is ROUGE \cite{lin2004rouge}. ROUGE evaluates n-gram co-occurrences between summary pairs. It works by comparing an automatically produced summary against a set of reference summaries.  The reference summaries are typically human produced, which are expensive and time-consuming. It is even more difficult when facing large amounts of texts in the big data age.  
There is no reference summary as golden standard in our selected dataset (i.e. SogouCA). To build a quantization standard, we propose an unsupervised evaluation metric. We consider that a faithful summary should:
\begin{enumerate}
\item Overlaps with title in three aspects: n-gram, domain knowledge keywords and named entities;
\item Contains more domain knowledge keywords than other non-summary texts;
\item Contains more named entities than other non-summary texts;
\item The similarities between two summary texts should be lower in case of redundancy.
\end{enumerate}

Formally, for a document $d$ in domain $D$, whose title is $t$, the faithful score of a generated summary $s$ is computed as:\begin{equation}
Score(s)=\mathbf{w} \cdot [ROUGE(t, s),  \frac{Count_{dkb}(t, s)}{Count_{dkb}(d)}, \frac{Count_{ent}(t, s)}{Count_{ent}(d)}, \frac{\sum_{e_1, e_2\in s} ROGUE(e_1, e_2)}{Count_{edu}(s)}, \frac{Count_{dkb}(s)}{Count_{dkb}(d)}, \frac{Count_{ent}(s)}{Count_{ent}(d)}]+b
\end{equation}
where $\mathbf{w}\in \mathbb{R}^N$, $b$ is a scalar. $ROUGE(a,b)$ denotes the ROUGE score between text $a$ and $b$. $Count_{dkb}(a,b)$ ($Count_{ent}(a,b)$) denotes the number of domain keywords (named entities) that $a$ and $b$ both have. $Count_{dkb}(x)$ ($Count_{ent}(x)$) denotes the number of domain keywords (named entities) that $x$ has. $Count_{edu}(s)$ denotes the number of EDUs in $s$. To make the score more objective, the hyper-parameters $[\mathbf{w}, b]$ were learned using linear regression on DUC2002\footnote{\url{https://www-nlpir.nist.gov/projects/duc/data/2002_data.html}} dataset. In the training step, the faithful score for each golden standard is set to 100.

\subsection{Results and Analysis}

For each generated RS-tree, we applied \autoref{alg_subroutine_ts} for summary generation. In what follows, we present the results using our method and our comparison to previous works. Since our model is unsupervised, we compare it with existing unsupervised single-document summarization methods. The baselines include:
\begin{itemize}
\item \textbf{Lead} selects the leading sentences in the document until length limit to form a summary, which is often used as an official baseline of DUC.
\item \textbf{TextRank} is a graph-based text summarization model. It represents the document as a graph in which sentences are nodes and the edges between two sentences are connected based on the similarity between them.
\item \textbf{ILP} is a text summarization technique which utilizes Integer Linear Program (ILP) for inference under a maximum coverage model.
\item \textbf{SummCoder} is an unsupervised framework for extracting sentences based on deep auto-encoders.
\end{itemize}

We generated four versions of summary (word length limit=100, 200 and the rate of text reduction=10\%, 20\%). 
\autoref{tb_eval_results} shows the faithful score of our method and baseline approaches. Our proposed framework outperforms many of the existing text summarizers on SogouCA dataset in terms of our proposed faithful score such as ILP, graph-based approaches.

\begin{table}
\caption{Evaluation results on SogouCA dataset}
   \centering
   \begin{tabular}{lllll}
   \toprule
   Approaches & 10\% & 20\% & 50 & 100 \\
   \midrule
    Lead & 62.9 & 72.8 & 63.2 & 73.6 \\
    TextRank & 65.2 & 76.6 & 66.1 & 75.4 \\
    ILP & 66.7 & 79.5 & 66.9 & 79.7 \\
    SummCoder & 68.8 & 82.3 & 69.0 & 82.5 \\
    Ours & 71.1 & 85.7 & 72.6 & 86.3 \\
   \bottomrule
\end{tabular}
\label{tb_eval_results}
\end{table}

The final summaries obtained from a sample SogouCA document by each summarizer (i.e. Lead, TextRank, ILP, SummCoder, and Ours) with word length limit 100 are shown in \autoref{tb_case_study}. From the summaries, it can be observed that the result generated from our method is more informative than other methods. Our result can summarize the results of others. The summary generated by ILP is similar to that generated by SummCoder but it is different from those generated by TextRank.

\begin{table}[!t]
\caption{Case study with summary length=100}
   \centering
   \begin{tabular}{p{\columnwidth}}
   \toprule
   \textbf{Title} \\
\begin{CJK}{UTF8}{gkai}
空客称争取以合作方式参与中国大飞机项目
\end{CJK} \\
Airbus said it is seeking to participate in China's large aircraft project in a cooperative manner.\\
   \midrule   
   \textbf{Lead} \\
\begin{CJK}{UTF8}{gkai}
新华网天津5月30日电 空中客车中国公司总裁博龙在天津接受新华社记者独家采访时说，空客正与中方合作伙伴商议，争取以合作方式参与中国大飞机项目。对于中国正在研发的大飞机项目，空客正与中方合作伙伴商议争取以合作方式参与该项目。
\end{CJK}\\
Xinhuanet Tianjin, May 30th -- Bolong, the president of Airbus China, said in an exclusive interview with Xinhua News Agency in Tianjin that Airbus is negotiating with Chinese partners to participate in China's large aircraft project in a cooperative manner. For the China's developing large aircraft project, Airbus is negotiating with Chinese partners to participate in the project in a cooperative manner.
   \\
   \midrule
    \textbf{TextRank}\\
\begin{CJK}{UTF8}{gkai}
空中客车中国有限公司企业资讯部提供的情况:早在1999年，空中客车公司与中国航空工业第一集团公司签署协议，计划分阶段向中国转让A320系列飞机机翼制造技术和生产线，目标是到2007年底使中国能够为空中客车在英国布劳顿和北威尔士的工厂制造A320系列飞机完整的机翼结构。
\end{CJK}\\
Airbus China Ltd. Corporate Information Department provided: As early as 1999, Airbus and China Aviation Industry First Group signed an agreement to transfer the A320 series aircraft wing manufacturing technology and production line to China in stages, with the goal of At the end of 2007, China was able to manufacture the complete wing structure of the A320 family of aircraft for Airbus' plants in Broughton and North Wales, England.
\\
   \midrule
    \textbf{ILP} \\
\begin{CJK}{UTF8}{gkai}
空客正与中方合作伙伴商议争取以合作方式参与该项目。中国作为世界航空市场增长最快的国家，已成为空中客车和波音全球的竞争焦点。通过该中心，中国已承担空中客车于2005年10月6日正式发起的、最新的A350飞机项目5\%的工作份额。
\end{CJK}\\
Airbus is negotiating with Chinese partners to participate in the project in a cooperative manner. As the fastest growing country in the world aviation market, China has become the focus of competition for Airbus and Boeing worldwide. Through the center, China has undertaken a 5\% share of the latest A350 aircraft project officially launched by Airbus on October 6, 2005.
\\
   \midrule
    \textbf{SummCoder} \\
\begin{CJK}{UTF8}{gkai}
空客正与中方合作伙伴商议争取以合作方式参与该项目。双方均不断加大在华采购、投资和技术合作，双方在最新机型上的重要零部件生产，也都有中国参与。博龙认为中国的大飞机之路困难而漫长，“空客花了近40年才取得今天的成就，现在我们拥有实力雄厚的工业基地。”
\end{CJK}\\
Airbus is negotiating with Chinese partners to participate in the project in a cooperative manner. Both parties have also participated in the production of important parts and components on the latest models. Bolong holds that China's big plane is difficult and long. ``Airbus has spent nearly 40 years to achieve today's achievements, and now we have a strong industrial base."
\\
   \midrule
   \textbf{Ours}\\
\begin{CJK}{UTF8}{gkai}
空客正与中方合作伙伴商议，争取以合作方式参与中国大飞机项目。中国作为世界航空市场增长最快的国家，已成为空中客车和波音全球的竞争焦点。双方均不断加大在华采购、投资和技术合作，双方在最新机型上的重要零部件生产，也都有中国参与。
\end{CJK}\\
Airbus is negotiating with Chinese partners to participate in China's large aircraft project in a cooperative manner. As the fastest growing country in the world aviation market, China has become the focus of competition for Airbus and Boeing worldwide. Both parties have continuously increased procurement, investment and technical cooperation in China. Both parties have also participated in the production of important parts and components on the latest models.
\\
   \bottomrule
\end{tabular}
\label{tb_case_study}
\end{table}

\section{Concluding Remarks}
\label{sec_conclusion}

In this paper, we proposed a novel unsupervised rhetorical parsing architecture for single-document extractive summarization. The proposed approach mainly contains three parts: domain knowledge base construction, Chinese-oriented rhetorical parsing and level-wise extractive summarization. To the best of our knowledge, this is the first study to adopt translation idea for rhetorical parsing. 

Firstly, we proposed a domain knowledge base construction model based on representation learning. The learned DKB can provide a panorama for a domain, which has two important roles for rhetorical parsing. One is discourse segmentation, and the other one is guiding rhetorical relation identification. In the unsupervised rhetorical parsing model, we leveraged the idea of translation, and designed a novel attention-based sequence-to-sequence model for rhetorical relation identification. Then the subroutine-based ATS model can accept different word length limit or summarization ratio and provide content-balanced results based on RS-tree. To evaluate our generated summary results in an unsupervised way, we presented a faithful score, whose hyper-parameters were learned on DUC2002 dataset. 

Directions for future work are many and varied. One of challenges left for the future is to further improve the performance of rhetorical parsing. Such as introducing attribute grammar into the deep neural model. Another important further work would be to utilize RS-tree for multi-document summarization.

\section*{Acknowledgments}
The authors would like to thank the developers of Pytorch \cite{paszke2017automatic}. This work was supported by the National Key Research and Development Program of China under grant 2016YFB1000902; and the National Natural Science Foundation of China (No. 61232015, 61472412, and 61621003).

\section*{References}

\bibliography{refs}

\end{document}